\documentclass[11pt]{article}

\usepackage{acl}

\usepackage{times}
\usepackage{latexsym}

\usepackage[T1]{fontenc}

\usepackage[utf8]{inputenc}
\usepackage{amsmath}
\usepackage{todonotes}
\usepackage{booktabs, multirow} 
\usepackage{soul}
\usepackage{xcolor,colortbl} 
\usepackage{changepage,threeparttable} 
\usepackage{microtype}

\usepackage{inconsolata}

\usepackage{graphicx}

\usepackage{newunicodechar}
\newcommand{\lowblock}{\rule[0.3ex]{1em}{0.15ex}}
\newunicodechar{▁}{\lowblock}
\title{Mask and You Shall Receive: Optimizing Masked \\ Language Modeling For Pretraining BabyLMs}

\author{Lukas Edman$^{1,2}$\qquad Alexander Fraser$^{1,2,3}$ \vspace{.2cm}\\ 
$^{1}$School of Computation, Information and Technology, TU Munich \\
$^{2}$Munich Center for Machine Learning \\
$^{3}$Munich Data Science Institute \\
\vspace{.1cm} {\tt \small lukas.edman@tum.de}
}

\begin{document}
\maketitle
\begin{abstract}

We describe our strategy for the 2025 edition of the BabyLM Challenge. Our main contribution is that of an improved form of Masked Language Modeling (MLM), which adapts the probabilities of the tokens masked according to the model's ability to predict them. The results show a substantial increase in performance on (Super)GLUE tasks over the standard MLM. We also incorporate sub-token embeddings, finding that this increases the model's morphological generalization capabilities. 
Our submission beats the baseline in the \texttt{strict-small} track.

\end{abstract}

\section{Introduction}

Traditionally, language models (LMs) have required billions to trillions of tokens for training, much less than a human typically sees, all while still suffering from the inability to accomplish relatively trivial tasks for humans. The 3rd BabyLM Challenge \cite{charpentier2025babylmturns3papers}, asks if we can train models more efficiently, making it more akin to the efficiency of human learning. 

A notable difference in standard LM training from human learning is that schooling is typically organized in curricula, meanwhile LMs tend to train on data in an unstructured, random manner. Thus, it is natural to consider bringing the concept of curriculum learning to LM training. Several works have attempted this, with only minimal success being shown in BabyLM's tracks \cite{warstadt-etal-2023-findings,hu-etal-2024-findings}. 

Our approach\footnote{\url{https://github.com/Leukas/babylm25}} for this year's BabyLM returns to this ever-elusive goal of effectively incorporating curriculum learning by optimizing the Masked Language Modeling (MLM) objective used to train encoder models. MLM by default masks every token with equal probability, but this is likely not optimal. Certain tokens that are easy to predict are likely a waste of time to mask, while other tokens that are more difficult to mask may require the model to learn key language concepts in order to reliably predict them. 

We introduce a form of MLM that adapts over the course of training, weighting the probabilities of masking individual tokens differently, based on the model's performance predicting them.

We also introduce an entirely different concept, designed instead to incorporate sub-token level information into the model's embeddings. Many works have shown the potential benefits of a model having access to sub-tokens or individual characters. While the evaluation tasks from previous BabyLM years did not require such finer-grained information to complete them, adjective nominalization was added as a task this year, alongside a similar task of converting to past tense, which was added as a hidden task for the final evaluation. We expect finer-grained character information to be useful for this task, especially.

\section{Related Work} \label{sect:related}
We focus on works related to our novel methods: Adaptive MLM and N-hot encodings.

\subsection{Masked Language Modeling}
A number of works have looked at improving the Masked Language Modeling objective. \citet{wettig-etal-2023-mask} experimented with the probability of masking a token, finding that values higher than the standard 15\% worked well in certain tests. \citet{yang-etal-2023-learning} continued along these lines and found that higher probabilities work better early in training, and lower values are better later. They also adjusted the probabilities of words being masked based on their POS tag, arguing that some word classes are much easier to predict and thus a waste of training time. \citet{belfathi2024languagemodeladaptationspecialized} similarly weigh words based on their domain specificity in order to do domain adaptation.

The most similar work to ours is \citet{zhang2023weightedsamplingmaskedlanguage}, whose dynamic masking strategy works similarly to our soft approach. They also weigh tokens based on their respective loss, but instead with the explicit purpose of oversampling rare tokens. This is used as a further pre-training strategy for BERT, and shows some limited improvement. 

In terms of cognitive plausibility, the adaptive method we introduce has some similarities to human behavior. In eye-tracking studies, humans tend to fixate on words that are more difficult to predict \cite{ehrlich1981contextual, rayner1996effects}. EEG studies have similarly shown that unpredictable words require more cognitive effort to process \cite{kutas1984brain}. Our method similarly steers the model to focus more on words that are difficult to predict.

\subsection{Character-level Information}
A number of works have sought to include character information in models. CharacterBERT \cite{el-boukkouri-etal-2020-characterbert}, ByT5 \cite{xue-etal-2022-byt5}, Byte Latent Transformer \cite{pagnoni-etal-2025-byte}, among many others have attempted to incorporate character or byte-level information within large-scale models. These works have noted that character-level models tend to train more efficiently, showing the normalized loss (bits-per-byte) can reach the same level in fewer steps, but they have not been extensively studied in a limited-resource setting such as BabyLM. 

For BabyLM itself, \citet{edman-bylinina-2023-much} have attempted to first pretrain on a character-level vocabulary and swapping to a BPE vocabulary without much success. \citet{goriely-etal-2024-babble} trained phoneme-level models but did not find improvements on the BabyLM benchmarks. The lack of improvements could be due to BabyLM not sufficiently measuring the models' understanding of orthography, phonology, or other aspects that require finer-grained information within the inputs. New to this year however is the adjective nominalization and past tense tasks \cite{doi:10.1073/pnas.2423232122}, which measure a model's morphological intuition by choosing the perceived correct ending to an imaginary adjective in order to convert it to a noun, or the perceived correct past-tense form of an imaginary infinitive verb.

\section{Method}
We first describe our adaptive masked language modeling (AMLM) scheme, then our token-level n-hot embedding architecture, and finally note the experimental details.

\subsection{Adaptive MLM}
The goal of AMLM is to improve the masking strategy such that we train the model on tokens from which it can learn the most. Therefore, tokens that are easy to predict should be assigned a lower probability of being masked. We employ 2 metrics to weigh each token: accuracy (\textbf{hard}) and loss (\textbf{soft}).

For both metrics, we start with a uniform probability for each token in the vocabulary:
\begin{equation}
w_{t=0,i} = p_{\mathrm{mlm}}, \quad \forall i \in V    
\end{equation}
where $p_{\mathrm{mlm}}$ is the overall probability of a token being masked in any given sequence, typically 15\%.
For every batch, we record the statistics of whether the model correctly predicted the masked tokens and the token-level loss of masked tokens. At the start of every timestep $t$ (which we define as 200 batches), we update the probabilities:
\begin{align}
&w_{t,i} = \lambda \, w_{t-1,i} + (1-\lambda) \, \tilde{w}_{t,i} \\
&\tilde{w}_{t,i} = p_{\mathrm{mlm}} \left( 1 - \mathrm{score_{t-1,i}} \right) 
\end{align}
where $\mathrm{score}_{t,i}$ is a scoring function, using either the hard or soft metric. We set $\lambda = 0.2$ empirically, which weighs the most recent statistics highly, but accumulates with previous timesteps nonetheless.
For the accuracy-based \textbf{hard} metric, our scoring function is a smoothed accuracy:
\begin{equation} \label{eq:score_smooth}
    \mathrm{score}_{t,i} = \dfrac{\mathrm{correct}_{t,i} + 0.5}{\mathrm{total}_{t,i} + 1} 
\end{equation}
where $\mathrm{correct}_{t,i}$ and $\mathrm{total}_{t,i}$ refer to the number of correctly predicted tokens of type $i$ at timestep $t$, and the total number of predicted tokens.
For the loss-based \textbf{soft} metric, the scoring function is a normalized, inverted loss:
\begin{equation}
    \mathrm{score}_{t,i} = 1 - \mathrm{norm}(\ell_{t,i})
\end{equation}
with $\ell_{t,i}$ corresponding to the average cross-entropy loss of token $i$ for timestep $t$. 

We allow the scores to range between 0 and 1, so that if the model is perfect at predicting a token (a score of 1), the probability of masking said token tends to 0. Meanwhile, a score of 0 causes the probability to tend towards $p_{\mathrm{mlm}}$. Finally, when masking each input sequence, the probabilities per token are normalized such that the average is $p_{\mathrm{mlm}}$, allowing individual tokens' mask probabilities to exceed $p_{\mathrm{mlm}}$.

\subsection{Token-level N-hot Embeddings}
Another strategy we try is incorporating more character-level information into the input embeddings. We accomplish this by what we call token-level n-hot embeddings, and it is best illustrated with an example: For the token \textit{\_doing}, we get all of the substrings that are also in our vocabulary, e.g., \textit{\_doin}, \textit{g}, \textit{\_do}, \textit{ing}, etc. These substrings are then encoded as an n-hot feature vector, i.e., 1 for \textit{\_doin}, \textit{g}, \textit{\_do}, \textit{ing}, etc., and 0 for everything else (hence the name n-hot). We then project this encoding into the embedding space with a linear layer, and add that to a separate, standard token embedding. To make this efficient, all of the n-hot encodings can be pre-calculated, with only the linear layer being trained on-the-fly, making the increase in training time negligible.

This strategy should be useful for any tasks that involve sub-token information. In particular, the adjective nominalization task asks for the model to provide the most plausible nominalization to a made-up adjective, e.g., ``wugable'' $\rightarrow$ ``wugability''. The ability for token-level n-hot encodings to trivially encode morphemes should make this task easier.

\subsection{Experimental Setup}
Our setup most closely follows that used by the strict-small GPT-BERT baseline\footnote{\url{https://huggingface.co/BabyLM-community/babylm-baseline-10m-gpt-bert-masked-focus}}, using the same learning rate, optimizer, and batch size. We use the same hidden and intermediate size for our model, however we use the Deberta-V2 \cite{he2021debertav3} architecture instead, given its ease of use and overall similarity to GPT-BERT, having the same attention mechanism.  We use a starting sequence length of 64 and raise it to 256 after 5 epochs. We use BPE with byte-fallback and a vocabulary size of 40k, following \citet{edman-bylinina-2023-much}'s findings that 40k appears near optimal. In terms of the probability of masking a token, we experiment with a decaying mask as suggested by \citet{yang-etal-2023-learning}, opting for 40\% at the start and linearly decaying to 15\%. We compare this to the standard constant 15\%. All of the hyperparameters are listed in Appendix \ref{app:hyperparams}.

In terms of data, we use the same data as in \citet{edman-etal-2024-babylms}, which consists of the initial BabyLM data, with the child-directed speech removed, and replaced with data from \citet{zhang-etal-2023-contrastive-learning}. Their data is synthetically generated triplets of sentences, paraphrases, and contradictions. We only use the data, not their contrastive learning approach. We compare this dataset to the original dataset from the shared task, as well as vocabulary size, in Appendix \ref{app:ablation}.

\begin{table*}[!htp]\centering
\small
\begin{tabular}{llrrrrrrrrr}\toprule
&&\multicolumn{3}{c}{Constant Mask} &\multicolumn{3}{c}{Decaying Mask} &\multicolumn{2}{c}{Hard + N-Hot} \\\cmidrule(lr){3-5} \cmidrule(lr){6-8} \cmidrule(lr){9-10}
&&Reg &Hard &Soft &Reg &Hard &Soft &Const & Decay \\\midrule
\multirow{10}{*}{Zero-shot}&BLiMP &70.7 &70.0 &69.7 &70.8 &\textbf{71.3} &71.4 &68.1 &67.7 \\
&Supplement &55.5 &56.9 &56.2 &57.8 &\textbf{58.3} &58.1 &56.2 &56.4 \\
&EWoK &50.6 &49.9 &50.7 &50.2 &\textbf{50.9} &50.1 &50.5 &50.2 \\
&Eye-tracking &9.0 &9.2 &\textbf{9.4} &9.1 &8.9 &8.8 &8.6 &8.5 \\
&Self-paced Reading &\textbf{4.2} &4.0 &4.0 &4.1 &3.7 &4.0 &3.9 &3.8 \\
&Entity Tracking &42.9 &43.8 &42.9 &\textbf{44.5} &44.4 &39.3 &43.3 &34.5 \\
&Adj Nominalization &35.3 &34.3 &22.0 &11.7 &14.3 &0.0 &\textbf{43.7} &42.0 \\
&Past-tense &4.0 &\textbf{6.3} &-6.7 &1.3 &1.3 &1.7 &-0.3 &4.0 \\
&COMPS &52.6 &53.1 &53.2 &53.9 &\textbf{54.0} &53.2 &52.0 &52.4 \\ \cmidrule(lr){2-10}
&Zero-shot Avg &36.1 &\textbf{36.4} &33.5 &33.7 &34.1 &32.2 &36.2 &35.5 \\ \midrule
\multirow{8}{*}{Finetune} &BoolQ &69.7 &68.6 &69.7 &\textbf{70.2} &69.2 &69.6 &67.3 &\textbf{70.2} \\
&MNLI &59.1 &59.9 &59.3 &61.8 &62.3 &61.6 &59.3 &\textbf{62.5} \\
&MRPC &88.1 &89.6 &88.2 &89.8 &\textbf{90.6} &89.8 &85.7 &89.1 \\
&QQP &72.5 &72.6 &72.4 &72.9 &\textbf{73.1} &72.9 &72.2 &72.8 \\
&MultiRC &68.1 &65.2 &\textbf{68.3} &64.6 &68.2 &67.9 &64.4 &68.4 \\
&RTE &57.8 &60.2 &58.0 &62.1 &\textbf{64.3} &62.6 &62.1 &60.2 \\
&WSC &64.1 &66.7 &64.1 &66.7 &63.5 &\textbf{67.3} &\textbf{67.3} &66.0 \\ \cmidrule(lr){2-10}
&Finetune Avg &68.5 &70.0 &68.6 &69.4 &\textbf{70.1} &69.9 &68.3 &69.9 \\ 
\bottomrule
\end{tabular}
\caption{Results of Regular MLM versus Hard and Soft AMLM, averaged over 3 runs. We also show performance when adding n-hot encodings using hard AMLM. }\label{tab:main}
\end{table*}

\section{Results}
We first present our results using our MLM objective. In Table \ref{tab:main}, we see that, overall, AMLM with the hard metric and decaying mask performs best overall. In general, the hard method performs better than the soft as well as regular MLM. While the differences are not very large, the hard mask strategy performs consistently better across multiple runs. 

In terms of using a decaying mask versus a constant one, the results are not as immediately clear, mainly due to adjective nominalization having a strong effect on zero-shot performance. In head-to-head comparisons for zero-shot, the decaying mask is better in 55\% (15/27) of tasks, tied in 7\% (2/27), and worse in 37\% (10/27). 
Combined with the better performance in fine-tuning, this confirms evidence from \citet{yang-etal-2023-learning}.

\subsection{N-hot Encodings}
We show the scores for the model with n-hot encodings also in Table \ref{tab:main}. We show only results with n-hot alongside the hard method, as it is the most performant, but the n-hot encodings are compatible with any form of pretraining.

On average, the n-hot encodings do not appear better or worse. Focusing on specific tasks, they perform noticeably worse on BLiMP. We suspect this is due to a sub-optimal manner of combining the n-hot embeddings with the regular ones. The performance on the (Super)GLUE tasks is comparable, suggesting that fine-tuning this model for a task may still yield competitive results. The performance on adjective nominalization is the most promising. There, we see the performance increase from 10 to 30 points over the comparable hard methods.  


It is not entirely clear why we do not see the same increase with the past-tense task. This may be due to the past-tense examples including more options, lowering the chance of human agreement, the options being more irregular, e.g. the past tense of ``veed'' could be ``veeded'', ``ved'', or ``vode'', or there being more collisions with real words, e.g. ``scor'' possibly becoming ``scored''. Ultimately, these tasks have no objectively correct answer. It would be better to measure the performance of n-hot encodings and similar methods that incorporate character-level information on tasks that require such information, like morphological inflection, for example.

\subsection{AMLM Analysis} \label{analysis}
As we record the statistics of the model's masked-token predictions, we can analyze the fluctuations of the probabilities with respect to various properties of words. We focus on two here: frequency and part-of-speech (POS).

\begin{figure}
    \centering
    \includegraphics[width=\linewidth]{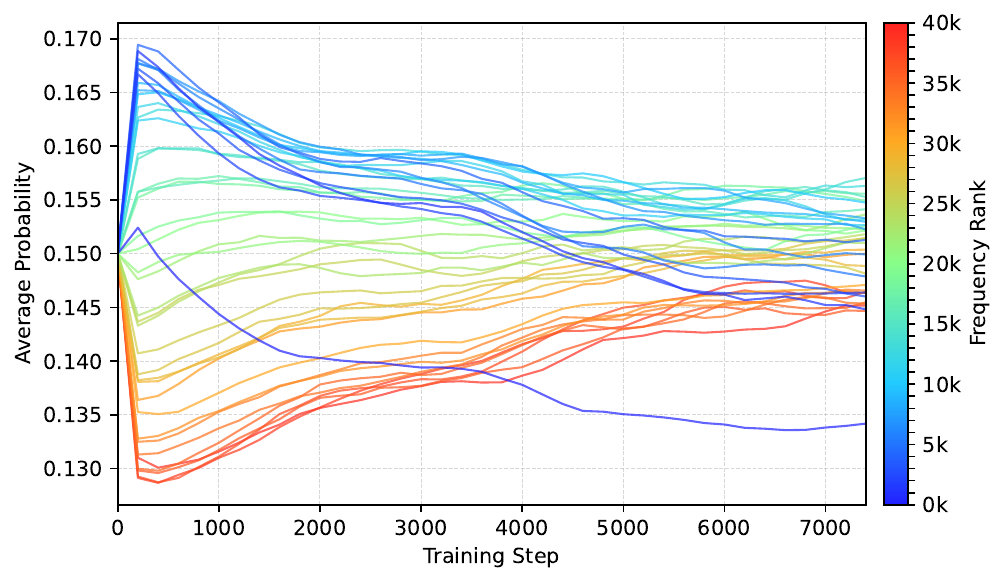}
    \caption{Token masking probabilities for hard method, grouped by frequency rank (in groups of 1000). Lower rank indicates higher frequency (e.g., blue is the most frequent group of words).}
    \label{fig:freq_hard}
\end{figure}

\begin{figure}
    \centering
    \includegraphics[width=\linewidth]{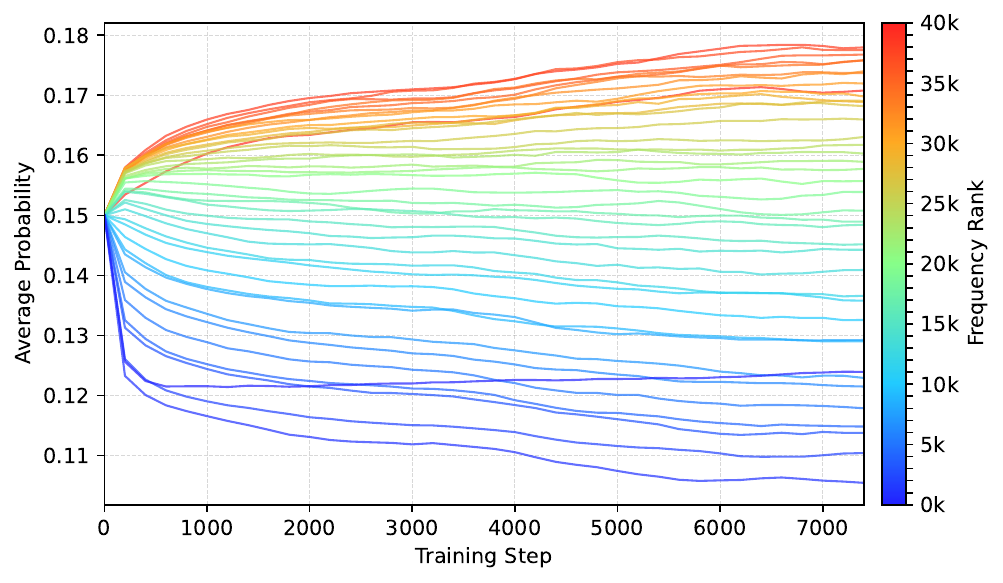}
    \caption{Token masking probabilities for soft method, grouped by frequency rank (in groups of 1000). Lower rank indicates higher frequency.}
    \label{fig:freq_soft}
\end{figure}

In Figure \ref{fig:freq_hard}, we can see the average probabilities of masking words using the hard method, grouped by frequency, in bins of 1000. Here, we see that initially, more frequent tokens are weighted higher, and rarer tokens lower. The weights of the top 5000 tokens quickly drop down, but the middle 20000 end up generally higher. The bottom 10000 tokens rise in weight but not back to 15\%. 

Figure \ref{fig:freq_soft} shows the soft masking probabilities, with a similar trend over time but quite different initial steps. The start sees common tokens quickly dropping in probability and rare words rising. The stark difference is due to the continuous nature of the soft method versus the discrete nature of the hard method. In the soft method, the loss goes down steadily, which immediately affects the probabilities. In the hard method, the accuracy does not immediately go up, as it takes time for the model to promote the correct token to the highest probability. The smoothing applied in Equation \ref{eq:score_smooth} causes the more common words to be favored for masking, hence the steep increase in the beginning.

Given that the hard method performs slightly better overall, these results indicate that masking common words more often at the start, even more often than the standard MLM does already due to their frequency in the text, may be a useful strategy for training more efficiently with MLM. As noted in Section \ref{sect:related}, our soft approach is similar to the strategy used by \citet{zhang2023weightedsamplingmaskedlanguage}, whose goal was to mask rarer tokens. Although the settings differ, our results may explain the limited improvement seen in their work.

\begin{figure}
    \centering
    \includegraphics[width=\linewidth]{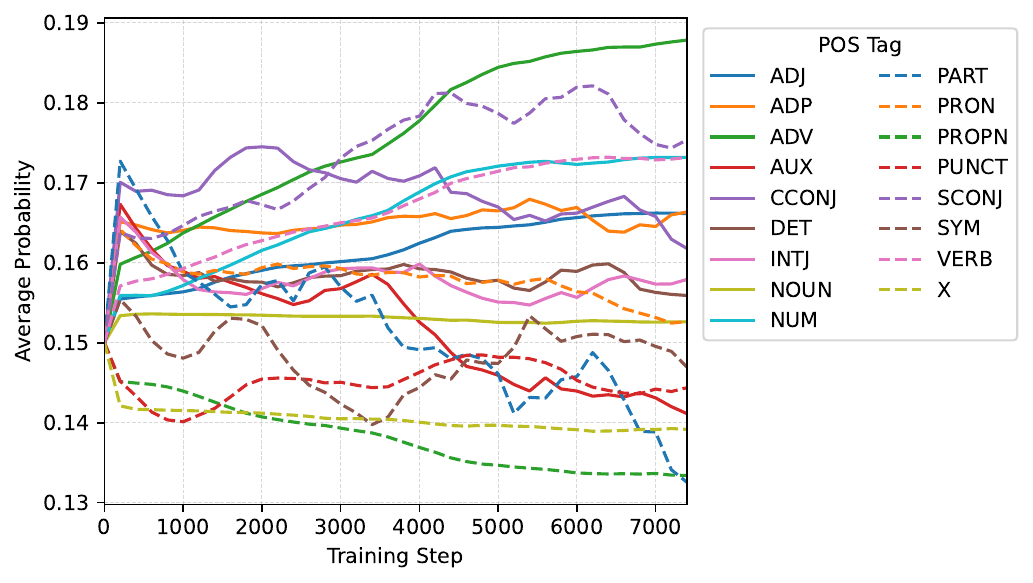}
    \caption{Masking probabilities by POS tag for hard method.}
    \label{fig:pos_hard}
\end{figure}

\begin{figure}
    \centering
    \includegraphics[width=\linewidth]{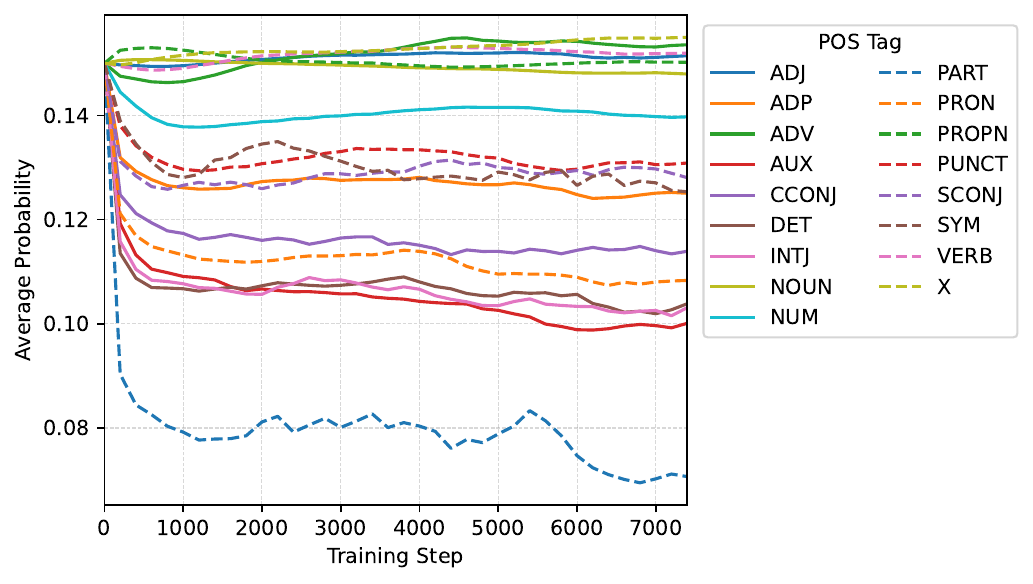}
    \caption{Masking probabilities by POS tag for soft method.}
    \label{fig:pos_soft}
\end{figure}

In Figure \ref{fig:pos_hard}, we can see the hard masking probabilities grouped by POS tag. We can see that the probabilities per POS tag are not uniform; instead, adverbs, subordinating conjunctions, numbers, and verbs are given higher weights. This tracks in general with intuition: adverbs have a lot of flexibility in their usage, as they can modify verbs, adjectives, or other adverbs, making them difficult to predict. Subordinating conjunctions being weighted higher might indicate that the model has difficulty understanding logical connections between multiple clauses. Exact numbers can also be difficult to predict.

Meanwhile, we see that particles, proper nouns, and X are weighted lower. The most common particle, ``to'', should be easy to predict when the model can see that the next token is an infinitive verb. Proper nouns may also be relatively easy, as their usage is usually very context-dependent. X comprises mainly of subtokens, e.g., ``arrog'', which would typically be part of ``arrogant'' or ``arrogance'', but is split by BPE due to the limited vocabulary size. 

In Figure \ref{fig:pos_soft}, we see a very different story: X and proper nouns are weighted highly while most other classes are weighted lower. It may seem as if the average does not equal 15\%, but X and PROPN, along with NOUN, are the 3 most common token classes in the vocabulary, making up 8.4k, 9.2k, and 12.4k, respectively.

The fact that we are POS tagging on the token level brings the unfortunate side effect of grouping several word parts into the X category. However, given that the X category, along with proper nouns, constitutes the majority of the difference, it is interesting to speculate why. Given their high weight for the soft method and low weight in the hard method, these words are often correctly predicted, but with a high loss. This is likely due to there being several candidate tokens that would fit the sentence. For example, ``arrog'' could easily be replaced with `ignor'' without much change in meaning. Further analysis is needed for a better understanding of this difference.

\begin{table*}[!htp]\centering
\small
\begin{tabular}{llrrrr}\toprule
& &\multicolumn{1}{c}{BertGPT} &\multicolumn{3}{c}{AMLM} \\\cmidrule(lr){3-3}\cmidrule{4-6}
& &Masked MNTP &Hard &Hard Decay &N-hot Hard \\\midrule
\multirow{11}{*}{Zero-shot} &BLiMP &70.4 &69.9 &\textbf{71.4} &65.6 \\
&Supplement &\textbf{63.7} &57.9 &59.2 &54.8 \\
&EWoK &50.0 &50.0 &\textbf{51.0} &49.7 \\
&Eye-tracking &\textbf{9.4} &8.9 &8.3 &8.1 \\
&Self-paced Reading &3.4 &\textbf{4.0} &3.5 &3.3 \\
&Entity Tracking &40.0 &43.6 &\textbf{44.2} &43.7 \\
&Adj Nominalization &2.7 &41.0 &34.0 &\textbf{64.0} \\
&Past-tense &\textbf{28.7} &8.0 &6.0 &1.0 \\
&COMPS &53.5 &52.3 &\textbf{54.2} &51.3 \\
&AoA &0.3 &-15.0 & -0.9 &\textbf{16.3} \\ \cmidrule(lr){2-6}
&Zero-shot Avg &32.2 &32.1 &33.1 &\textbf{35.9} \\ \midrule
\multirow{8}{*}{Finetune} &BoolQ &67.6 &69.2 &\textbf{69.5} &68.9 \\
&MNLI &51.4 &59.3 &\textbf{62.3} &58.5 \\
&MRPC &86.1 &\textbf{90.3} &\textbf{90.5} &81.0 \\
&QQP &67.4 &72.4 &\textbf{73.1} &72.2 \\
&MultiRC &\textbf{71.6} &69.8 &68.6 &57.5 \\
&RTE &57.5 &61.1 &\textbf{63.3} &64.0 \\
&WSC &61.5 &\textbf{73.1} &63.5 &69.2 \\\cmidrule(lr){2-6}
&Finetune Avg &66.2 &\textbf{70.7} &70.1 &67.3 \\\midrule 
&Final Score &38.2 &34.2 &38.3 &\textbf{41.9} \\ 
\bottomrule
\end{tabular}
\caption{Our submission models (AMLM) versus the comparable BabyLM baseline (BertGPT). Final score refers to the scoring equation used by the BabyLM evaluation leaderboard.}\label{tab:submission}
\end{table*}

\subsection{Submission}
Here we compare our submitted models to the leaderboard with the strongest baseline provided by the organizers. We select only models trained with RNG seed 0 so as not to overfit to the evaluation benchmark. The results are shown in Table \ref{tab:submission}.

Here, we can see that our hard decay model and hard n-hot models outperform the baseline according to the aggregate score. For the hard decay model, the main improvement appears to be the (Super)GLUE scores, however these are aggregated into one value for the final score, so their weighting is less important. Entity tracking seems to be the only other culprit, as the rest of the scores are fairly close to each other.

The n-hot model stands out in its performance on adjective nominalization, which is enough for the average on zero-shot tasks to favor this model, as well as the aggregate score. As the n-hot model performs poorly on most other tasks, the final scoring metric seems to be unfairly skewed. We further question the efficacy of the BabyLM metrics in Appendix \ref{app:metrics}, where we find that a model whose training loss spikes due to a mistake on our part actually gets a higher aggregate score than any model introduced so far. 



\section{Conclusion}
The BabyLM Challenge challenges us to train language models on a limited data budget and, for the first time this year, a limited training time budget (by way of epochs). We show that greater training efficiency can be achieved through Adaptive MLM, which changes the probabilities of tokens being masked during training, according to their difficulty. The results show an increase in performance, beating the baseline set by the organizers.
We also investigate a method of incorporating subtoken-level information into the model, which showed promising performance on adjective nominalization, the task that requires finer-grained, morpheme-level understanding.

There is plenty of future work to be investigated. The adaptive masking scheme required setting several hyperparameters empirically, and is likely far from optimized. While we use generic statistics based on individual tokens, groups of tokens being masked in tandem may be more challenging and force the model to learn more properties of language. For this, we would suggest word-level or phrase-level masking, or more interestingly, a neural masker, which learns to mask tokens based on what is most challenging for the main model to predict. 

Our subtoken approach is also far from optimal. In theory, it should be possible to incorporate two or more granularities of input without any negative effects on downstream performance. Such improvements would have a considerable impact not only in the BabyLM sphere, but higher-resource NLP as well.

\section*{Acknowledgements}
We thank BabyLM anonymous reviewers for the helpful comments. The work was supported by the European Research Council (ERC) under the European Union's Horizon Europe research and innovation programme (grant agreement No. 101113091) and by the German Research Foundation (DFG; grant FR 2829/7-1).

\bibliography{acl_min,custom}

\appendix

\begin{table*}[!htp]\centering
\small
\begin{tabular}{lrrrrrr}\toprule
&\multicolumn{2}{c}{BertGPT Baselines} &\multicolumn{2}{c}{AMLM - Hard Decay} & \\\cmidrule(lr){2-3} \cmidrule(lr){4-5}
&Causal &Masked &Success &Failure &Untrained \\\midrule
BLiMP &\textbf{71.7} &70.4 &71.4 &59.0 &48.8 \\
BLiMP Supplement &63.2 &\textbf{63.7} &59.2 &51.7 &42.3 \\
EWoK &49.5 &50.0 &51.0 &\textbf{56.0} &50.0 \\
Entity Tracking &34.6 &40.1 &\textbf{44.2} &41.3 &41.7 \\
Adj Nominalization &59.2 &2.7 &22.3 &\textbf{78.1} &77.4 \\
Past Tense &12.9 &\textbf{28.7} &6.2 &-12.1 &-16.2 \\
COMPS &52.8 &53.6 &54.2 &\textbf{82.2} &50.0 \\
Reading &\textbf{6.7} &6.4 &5.9 &6.4 &5.7 \\
AoA &-3.9 &0.3 &-0.9 &\textbf{34.2} &0.0 \\
(Super)GLUE &65.1 &66.0 &\textbf{69.8} &57.7 &62.1 \\ \midrule
Avg &41.2 &38.2 &38.3 &\textbf{45.4} &36.2 \\
\bottomrule
\end{tabular}
\caption{Baselines and our best submitted model, compared to a failed run and untrained model.}\label{tab:wtf}

\end{table*}

\section{Training Hyperparameters} \label{app:hyperparams}

We note the complete hyperparameters used in Table \ref{tab:hyperparams}. 

\begin{table}[!htp]\centering
\begin{tabular}{lrrr}\toprule
&Parameter &Value \\\midrule
\multirow{5}{*}{Model} &Architecture &Deberta \\
&Hidden Size &384 \\
&Intermediate Size &1280 \\
&Dropout &0.1 \\
&Vocabulary Size &40000 \\ \midrule
\multirow{14}{*}{Training} &Sequence Length &64, 256 \\
&Batch Size (in tokens) &16384 \\
&Learning Rate &7e-3 \\
&Epochs & 10 \\
&Number of Steps &8325 \\
&Scheduler & Cosine \\
&Warmup Ratio &1\% \\
&Mask Ratio &0.4→0.15 \\
&Random Ratio &0.1 \\
&Keep Ratio &0.1 \\
&Weight Decay &0.01 \\
&Optimizer &LAMB \\
&Optimizer Epsilon &1e-8 \\
&Optimizer Beta 1 &0.9 \\
&Optimizer Beta 2 &0.95 \\
&Grdient Clipping &1 \\
\bottomrule
\end{tabular}
\caption{Hyperparameters used (except where otherwise indicated).}\label{tab:hyperparams}
\end{table}

\section{Ablation of Dataset and Vocabulary Size} \label{app:ablation}
Two of the major differences between our models and the baselines trained by the organizers are the training dataset and vocabulary size. We ablate the two with a standard MLM training scheme and our hard AMLM method in Table \ref{tab:ablate}.

\begin{table*}[!htp]\centering
\begin{tabular}{llrrrrrrrrr}\toprule
& &\multicolumn{4}{c}{Regular} &\multicolumn{4}{c}{AMLM - Hard} \\\cmidrule(lr){3-6} \cmidrule(lr){7-10}
& &\multicolumn{2}{c}{Original} &\multicolumn{2}{c}{New} &\multicolumn{2}{c}{Original} &\multicolumn{2}{c}{New} \\\cmidrule(lr){3-4} \cmidrule(lr){5-6} \cmidrule(lr){7-8} \cmidrule(lr){9-10}
& &16k &40k &16k &40k &16k &40k &16k &40k \\\midrule
\multirow{10}{*}{Zero-shot} &BLiMP &66.9 &66.4 &67.2 &70.7 &67.5 &67.0 &68.8 &70.0 \\
&Supplement &58.9 &59.2 &58.3 &55.5 &62.0 &59.0 &57.9 &56.9 \\
&EWoK &49.9 &50.6 &49.0 &50.6 &51.2 &50.8 &50.3 &49.9 \\
&Eye-tracking &6.2 &6.3 &7.2 &9.0 &6.5 &5.9 &6.7 &9.2 \\
&Self-paced Reading &3.6 &3.2 &4.0 &4.2 &3.5 &4.0 &3.6 &4.0 \\
&Entity Tracking &31.2 &27.5 &32.6 &42.9 &32.1 &35.6 &36.9 &43.8 \\
&Adj Nominalization &37.0 &18.0 &56.0 &35.3 &50.0 &11.0 &43.0 &34.3 \\
&Past-tense &29.0 &18.0 &0.0 &4.0 &23.0 &-10.0 &18.0 &6.3 \\
&COMPS &51.4 &52.0 &51.9 &52.6 &51.9 &51.9 &52.7 &53.1 \\ \cmidrule(lr){2-10}
&Zero-shot Avg &37.1 &33.5 &36.2 &36.1 &38.6 &30.6 &37.5 &36.4 \\ \midrule
\multirow{8}{*}{Finetune} &BoolQ &68.5 &67.8 &70.9 &69.7 &70.2 &68.4 &68.8 &68.6 \\
&MNLI &44.0 &47.3 &56.6 &59.1 &44.5 &48.5 &57.9 &59.9 \\
&MRPC &82.6 &82.0 &86.7 &88.1 &82.7 &83.7 &88.7 &89.6 \\
&QQP &66.3 &67.8 &71.6 &72.5 &67.0 &68.8 &72.7 &72.6 \\
&MultiRC &67.1 &66.6 &66.6 &68.1 &66.2 &64.4 &67.8 &65.2 \\
&RTE &59.7 &59.0 &59.0 &57.8 &56.8 &53.2 &61.1 &60.2 \\
&WSC &65.4 &65.4 &61.5 &64.1 &63.5 &61.5 &65.4 &66.7 \\ \cmidrule(lr){2-10}
&Finetune Avg &64.8 &65.1 &67.5 &68.5 &64.4 &64.1 &68.9 &69.0 \\
\bottomrule
\end{tabular}
\caption{Ablation of vocabulary size and dataset.}\label{tab:ablate}
\end{table*}

First, concerning the dataset, the new dataset appears generally better for the finetuning tasks. Most of this increase is in the MNLI, MRPC, and QQP tasks. This aligns with the findings of \citet{edman-bylinina-2023-much}, and can be explained by the additional data having similar content (i.e., paraphrases). 

As for vocabulary size, the results appear dependent on the dataset. For the new dataset, the higher vocabulary size appears to perform better, especially for BLiMP and Entity Tracking. The higher vocabulary size may be slightly more optimal for the new dataset, but for the original dataset, the lower vocabulary size appears more optimal. The difference in zero-shot performance on adjective nominalization and past-tense is quite substantial, though the variance in these metrics is high, so it is difficult to gauge the importance of these results.

\section{Should we trust BabyLM Metrics?} \label{app:metrics}
During our experimentation, we encountered some surprising results with respect to some of the metrics used to evaluate models. First, we found that one could obtain quite high scores on adjective nominalization, around 78, just from initialization, no training required. 

Furthermore, we accidentally trained a model with too low a batch size for the corresponding learning rate, causing the loss to spike during training and never recover. While it performs expectedly poorly on some metrics, such as BLiMP and (Super)GLUE, it performs remarkably well on others, namely EWoK, adjective nominalization (again), COMPS, and AoA. We show the results in Table \ref{tab:wtf}. 

This raises the question: should we trust the metrics used in BabyLM? For BLiMP and (Super)GLUE the answer appears to be yes. Failed and untrained models perform expectedly poorly on these. For self-paced reading and eye-tracking, the numbers appear to stay relatively similar, regardless of the model. And for the rest, their scores should probably be taken with a baby fist of salt.

\end{document}